\documentclass[letterpaper, 10 pt, conference]{ieeeconf}  

\IEEEoverridecommandlockouts                              

\usepackage{subcaption}
\usepackage{bm}
\usepackage[pdftex]{graphicx}
\usepackage{graphics} 
\usepackage{epsfig} 
\usepackage{mathptmx} 
\usepackage{times} 
\usepackage{amsmath} 
\usepackage{amssymb}  
\usepackage[noend]{algpseudocode}
\usepackage{multirow}
\usepackage{tikz}
\usepackage{lettrine}
\usepackage{color}
\usepackage[pagebackref=true,breaklinks=true,letterpaper=true,colorlinks,bookmarks=false]{hyperref}

\newboolean{highlight}
\setboolean{highlight}{false}
\colorlet{Mycolor1}{green!10!orange!90!}
\newcommand{\change}[1]{\ifthenelse{\boolean{highlight}}{\textcolor{red}{#1}}{#1}}

\newboolean{draft-mode}
\setboolean{draft-mode}{false}
\newcommand{\sidenote}[1]{\ifthenelse{\boolean{draft-mode}}{\marginpar{\tiny\raggedright\textsf{\hspace{0pt}#1}}}{}}
\DeclareRobustCommand{\arnote}[1]{\ifthenelse{\boolean{draft-mode}}{\textcolor{blue}{\textbf{AR: #1}}}{}}
\DeclareRobustCommand{\dmnote}[1]{\ifthenelse{\boolean{draft-mode}}{\textcolor{cyan}{\textbf{DM: #1}}}{}}

\newcommand{\secref}[1]{Section~\ref{#1}}

\newcommand{\figref}[1]{Fig.~\ref{#1}}
\newcommand{\myparagraph}[1]{\vspace{0.07in}\noindent\textbf{#1}}

\usepackage[ruled, lined, linesnumbered, commentsnumbered, longend]{algorithm2e}
\makeatletter
\newcommand{\removelatexerror}{\let\@latex@error\@gobble}
\makeatother
\usepackage{cuted}
\usepackage{capt-of}

\title{\LARGE \bf
Extrinsic Contact Sensing with Relative-Motion Tracking from Distributed Tactile Measurements
}

\author{Daolin Ma$^{1}$, Siyuan Dong$^{1}$ and Alberto Rodriguez$^{1}$
\thanks{*This research is supported by the HKSAR Innovation and Technology Fund (ITF) ITS-104-19F.}
\thanks{$^{1}$Mechanical Engineering Department, Massachusetts Institute of Technology, 77 Massachusetts Institute of Technology, MA 02139, US
        {\tt\small  <daolinma,sydong,albertor>@mit.edu}}%
}

\begin{document}

\maketitle
\thispagestyle{empty}
\pagestyle{empty}

\begin{abstract}

This paper addresses the localization of contacts of an unknown grasped rigid object with its environment, i.e., extrinsic to the robot. We explore the key role that distributed tactile sensing plays in localizing contacts external to the robot, in contrast to the role that aggregated force/torque measurements traditionally play in localizing contacts on the robot. When in contact with the environment, an object will move in accordance with the kinematic and possibly frictional constraints imposed by that contact. Small motions of the object, which are observable with tactile sensors, indirectly encode those constraints and the geometry that defines them.

We formulate the extrinsic contact sensing problem as a constraint-based estimation problem. The estimation is subject to the kinematic constraints imposed by the tactile measurements of object motion, as well as the kinematic (e.g., non-penetration) and possibly frictional (e.g., sticking) constraints imposed by rigid-body mechanics. We validate the approach in simulation and with real experiments on the case studies of fixed point and line contacts.

This paper discusses the theoretical basis for the value of distributed tactile sensing in contrast to aggregated force/torque measurements. It also provides an estimation framework for localizing environmental contacts with potential impact in contact-rich manipulation scenarios such as assembling or packing. 

\end{abstract}

\section{Introduction}
\label{sec:introduction}

\begin{figure}[t]
\centering
  \includegraphics[width=0.75\linewidth]{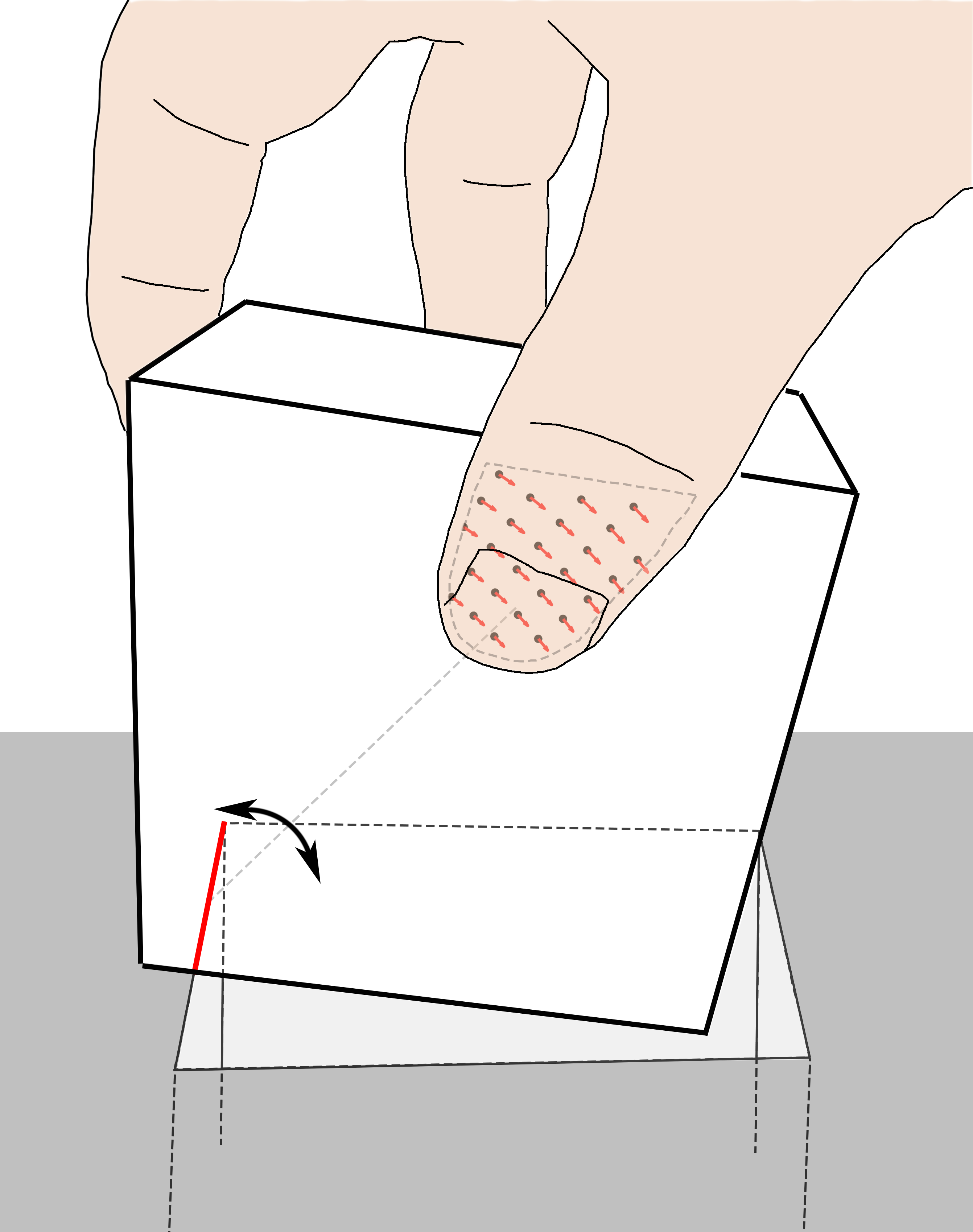}
  \caption{Localizing extrinsic contact while inserting an object.}
  \label{fig:inserting_box}
\end{figure}

\begin{figure*}[ht]
\centering
  \vspace{-4mm}
  \includegraphics[width=0.75\linewidth]{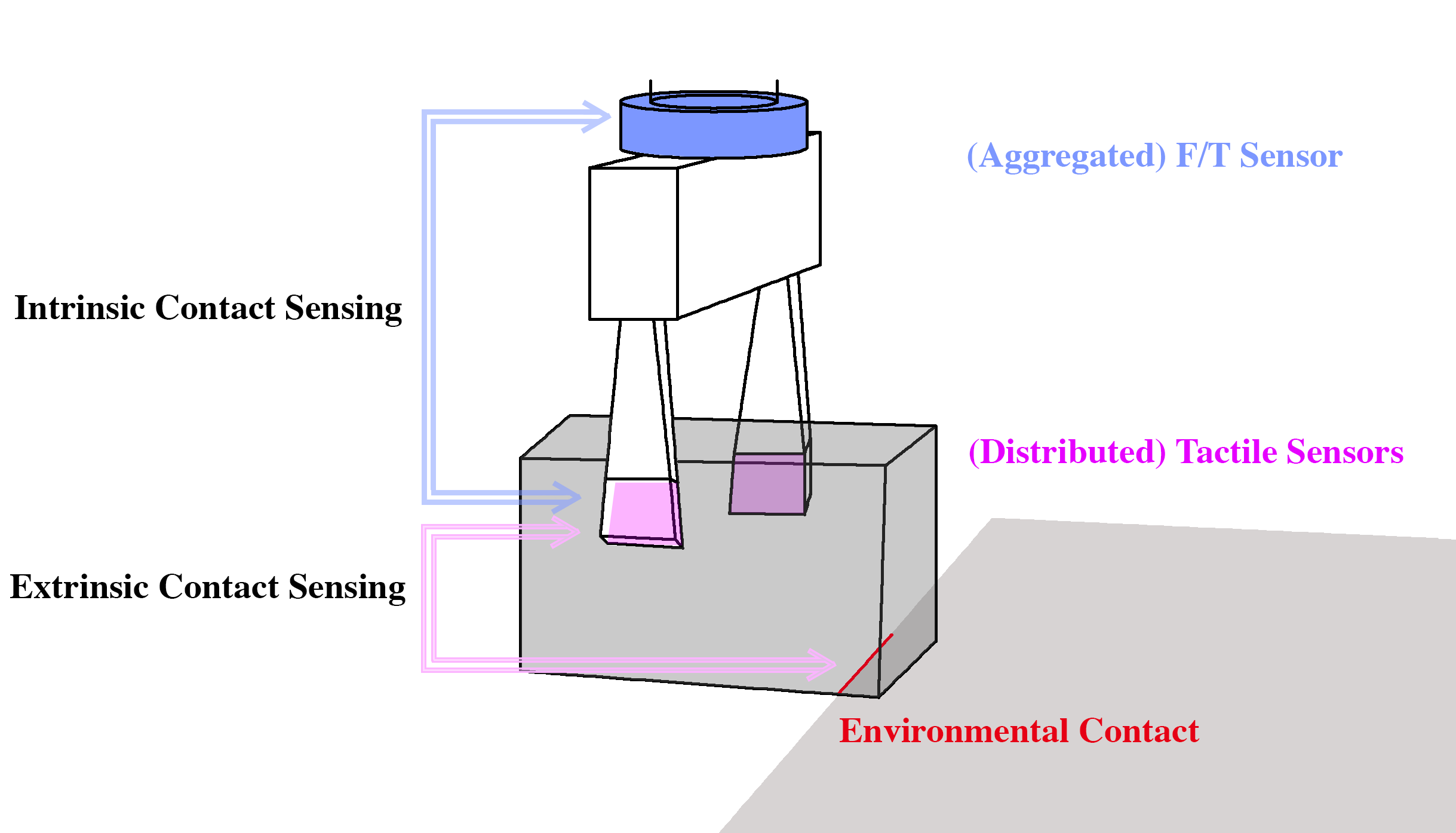}
  \caption{
      \textbf{Extrinsic contact sensing} estimates the contact between the object and environment using distributed {motion} sensors on robot; \textbf{Intrinsic contact sensing} estimates the contact between finger and the object using aggregated {Force/Torque} sensor on robot.
  }
  \label{fig:Extrinsic_vs_Intrinsic}
\end{figure*}

In this work we address the problem of localizing contacts between a grasped object of unknown geometry and its environment, which we refer to as \emph{extrinsic contact sensing}. 
The ability to infer the location of external contacts, which in the absence of environment instrumentation cannot be directly sensed, is of essence in insertion tasks (\figref{fig:inserting_box}), and is relevant to many other contact-rich manipulation tasks that involve using grasped objects (e.g., tools).

This is in contrast to \emph{intrinsic contact sensing}, introduced by Bicchi et al.~\cite{bicchi1993contact}, which addresses the problem of localizing contacts on the surface of a robot with known geometry.
A key aspect of this work is that we focus on the role that distributed tactile measurements play in localizing external contacts, in contrast to the role that aggregated force/torque measurements play in localizing internal contacts. 
Figure~\ref{fig:Extrinsic_vs_Intrinsic} illustrates the difference between the two approaches.

\myparagraph{Intrinsic Contact Sensing.} The goal is to build a map between resultant forces captured by a F/T sensor and the location of the contact force responsible for them on the surface of the robot body attached to that sensor. If the geometry of the body is known, the key challenge is to characterize when the map between contact forces and sensed wrist forces can be inverted.
Initial works were based on the realization that if the body attached to the sensor has a calibrated geometry, and is convex, and the force is at a point contact, that map is invertible. In practice, in the scenario in Fig.~\ref{fig:Extrinsic_vs_Intrinsic} this would require precise knowledge of the geometry of the wrist$\rightarrow$gripper$\rightarrow$fingers chain.
This idea can be extended to more complex estimation frameworks, for example to estimate contacts on a kinematic chain of known geometry~\cite{Manuelli2016}, or when there is compliance in between the body and the sensor~\cite{Yu2018}. 

\myparagraph{Extrinsic Contact Sensing.} If we add the grasped object to the manipulation chain wrist$\rightarrow$gripper$\rightarrow$fingers$\rightarrow$object, the assumption of known geometry becomes more impractical. We often do not have precise knowledge of the shape of the manipulated object, and the pose in the grasp is easily subject to change.
Instead, we address extrinsic contact sensing by exploiting the ability of tactile sensors to observe object motion. 

When in contact with the environment, an object will move in accordance with the kinematic and possibly frictional constraints imposed by that contact. Small motions of the object, when observable with tactile sensors, indirectly encode those constraints, and the geometry that defines them.
The approach we propose infers contact location from observations of that motion, even without requiring knowledge of the geometry of the object.
Object motion, in place, can be observed by tactile sensors that track contact deformation produced by the external contacts. We do this by exploiting tactile feedback, in our case, captured by the recently developed sensor GelSlim 3.0 \cite{taylor2021iros}, a high-spatial resolution tactile sensor that has been shown to be able to recover the strain and slip fields at contact~\cite{Elliott2018GelSlim, ma2019dense,dong2019maintaining}.

We formulate the extrinsic contact sensing problem in \secref{sec:approach} as a constraint-based estimation framework subject to the kinematic constraints imposed by the tactile measurements of object motion, as well as the kinematic (e.g., non-penetration) and possibly frictional (e.g., sticking) constraints imposed by rigid-body mechanics. A least-squares solver optimizes the constraints to find the location of an external contact directly from tactile observations.
We validate the approach with simulation and real experiments on the case study of point contact and an edge contact, detailed in \secref{sec:case_study}.

The proposed framework can also be generalized to different distributed sensor setups beyond high-resolution tactile sensors; for example, more sparser tactile sensors distributed through multiple finger contacts.

\subsection{Related Work}
Here we review some recent work that uses high-resolution vision-based tactile sensors to observe geometry, force, or slip.

Li \textit{et al.}~\cite{Li2014} presented the use of the tactile sensor GelSight, to track pose of small parts held in a robot hand with high accuracy. The researchers demonstrated that the accuracy was enough to localize a USB connector and insert it into the mating hole.
Izatt \textit{et al.}~\cite{Izatt2017} extended the approach and presented a filtering algorithm fusing vision and touch to track pose of an object held by treating the output of GelSight as a pointcloud. They demonstrated that the accuracy was sufficient to grasp a screwdriver and insert it in a hole. 
Both works assumed known object geometry and focused on model-based pose tracking. 

In recent work, instead, Dong \textit{et al.}~\cite{dong2019tactilebased} used a self-supervised data-driven approach to estimate the contact formation and contact location of a grasped object against external features, directly using GelSlim~\cite{Elliott2018GelSlim} high-resolution tactile imprints. 
This success in localizing contact using GelSlim without direct knowledge of the geometry of the object is a direct inspiration for the work in this paper. 
We'd like to answer questions \textit {"
What is the mechanism by which tactile sensors can recover information about external contact location?"} and  \textit{"What are the differences between using distributed tactile sensor measurements versus aggregated force/torque measurements?"}

\begin{figure*}[htb]
\centering
  \includegraphics[width=\linewidth]{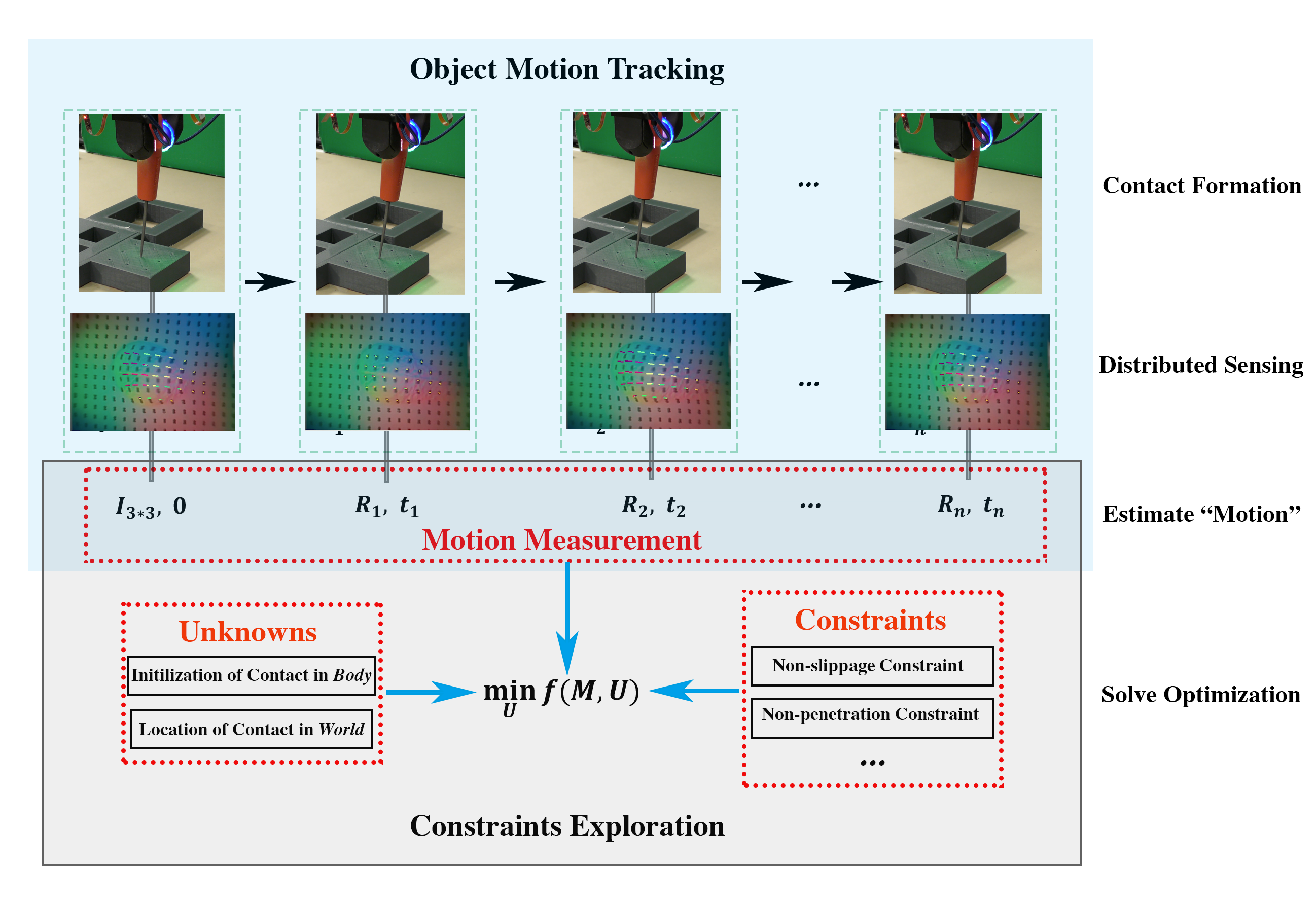}
  \vspace{-5mm}
  \caption{Framework of \textit{extrinsic contact localization} from distributed tactile measurements.}
  
  \label{fig:pipeline of contact localization}
\end{figure*}

\section{Framework for Extrinsic Contact Sensing}
\label{sec:approach}

The key idea is to recover the kinematic constraints that define the object's motion, which are captured by the tactile measurements.

\subsection{Assumptions}

We make the following assumptions to facilitate the formulation of this problem:

\begin{itemize}
    \item The grasped object is rigid.
    \item The environment is rigid.
    \item The object remains in contact with environment.
    \item The grasp of the object is stable.
\end{itemize}

First, both the object and environment are rigid and they remain in contact with each other, so that consistent constraint will be recoverable from object motion.
Second, the grasp of the object is stable so that no slip between object and the finger exists, which makes it easier to track from distributed tactile measurements. 

Many of these assumptions can be simplified by considering more complex (possibly learned) models of interaction, but in this paper we would like to focus on the fundamentals of the relationship between external contacts and haptic signals.

\subsection{Problem Formulation}

We formulate the estimation problem as a constraint satisfaction optimization problem. \figref{fig:pipeline of contact localization} illustrates the framework of contact localization via Constraint Optimization Problem and Object Motion Tracking. The constraints come in two forms: 1) Direct contact with the environment, and 2) observations of motion of the object, which we will describe in sections \secref{sec:Constraints} and
\secref{sec:Motion Tracking}.

By assumption, the object is rigid, it has 6 DOFs. Once in contact with the environment, its mobility is reduced by the constrains formed by contact. Assuming that the motion of the grasped object is track-able, we will see that motion measurements can be used to localize extrinsic contacts by exploring which and how certain DOFs are constrained.  

\subsection{Contact Kinematics Constraints}
\label{sec:Constraints}

    

We can add as many constraints as we have based on prior knowledge of the object and environment. These constraint equations embed the unknown parameters for contact location in environment, such as position of a point $\bm {P_w}$, direction of a line $\bm {e_w} $, or normal direction of a plane $\bm{n_w}$. These constraints can also embed the unknown parameters for localizing the contact point on the object, such as position of a point $\bm {P_b}$, direction of a line $\bm {e_b} $ , or normal direction of a plane $\bm{n_b}$. We denote the chosen unknowns as $\bm{U}$ based on the contact type, where 

$$\bm{U \subseteq \{P_w,e_w,n_w,P_b,e_b,n_b,...\}}.$$ 

For example, a fixed point contact, namely a point that remains in contact with the environment over a period of time, indicates that the position of this point $\bm {P_b}$ on object is constant in world frame even though the object might be experiencing a rigid rotation $\bm {R}(t)$  and translation $\bm {T}(t)$. This constraint can be expressed in the world frame as  
\begin{equation}
    (\bm {R}(t) \bm {P_b} + \bm {T}(t)) - \bm {P_b} = 0
    \label{eq:fixed_point_constraint}
\end{equation} 

Similarly, a fixed direction contact, namely a direction remains unchanged all through a period of  time, indicates that the direction  $\bm {e_b}$ on the object is constant even though the object might be experiencing a rigid rotation $\bm {R(t)}$  and translation $\bm {T(t)}$. This constraint can be expressed in the world frame as 
\begin{equation} 
(\bm {R}(t)-\bm{I}(3)) \bm {e_b} = 0    \label{eq:fixed_direction_constraint}
\end{equation}

An optimization can be formulated based on satisfying these constraints. For example, by minimizing $\bm{J_1} = ||(\bm {R}(t) \bm {P_b} + \bm {T}(t)) - \bm {P_b} ||$ or $\bm{J_2} = ||(\bm {R}(t)-\bm{I}(3)) \bm {e_b} ||$, where $\bm{I}(3)$ represents a 3$\times$3 identity matrix.

Therefore, the tracked rigid body motion is used to regress the unknown localization of the contacts. Given the measurement of object's motion $\bm{M(t) = \{ \bm{R}(t),\bm{T}(t)\}}$ over a time span, an optimization problem can be formulated as
\begin{equation}
    \min_{\bm{U}} {\bm{f(M, U)}},
\end{equation}
where $\bm{f(M, U)}$ corresponds to the residues of the implemented constraint equations. 

The above-mentioned fix-point or fix-direction constraints don't require knowledge of the object's geometry other than its relative rigid body motion in the world frame. It means that sticking point or line contact between an unknown object and the environment can be localized using this method. On the other hand, to formulate slipping non-penetration constraints, a minimum knowledge of local geometry around the contact pair is needed.

It's also worth noting that when assuming that the object is always in contact, the typical non-equality constraints for non-penetration are indeed equality constraints. 
Note also that the constraint equations can be expressed in either configuration space or velocity space. 
But measurement of object's motion in velocity space can be more noisy.

\subsection{Relative-motion Tracking  with Distributed Tactile Sensing}
\label{sec:Motion Tracking}

We track the relative motion $\bm{M}$ as an input to the optimization problem, with the high-resolution vision-based tactile sensor GelSlim-3.0. Key to such capability is that we can reconstruct the 3D positions of an array of markers in the contact patch, and the motions of these markers reflect the motion of the grasped object in the gripper frame. The motion of the grasped object in the world frame can be achieved by further coupling with the motion of the gripper in the world frame typically tracked via the built-in encoders of the robot-arm.
\begin{figure}[t]
\centering
  \begin{subfigure}{0.23\textwidth}
  \includegraphics[width=\linewidth]{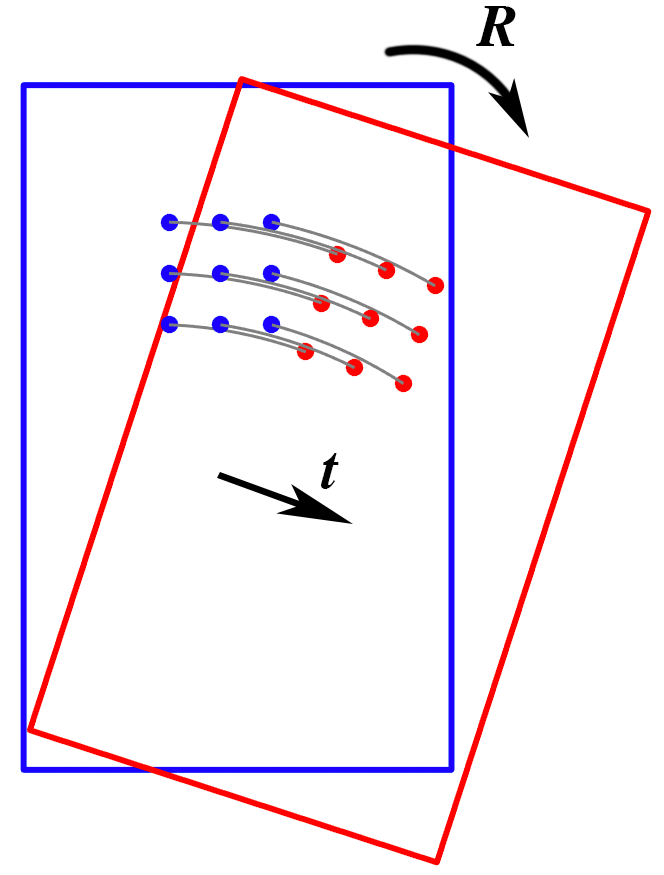}
  \caption{}
  \end{subfigure}
  \begin{subfigure}{0.23\textwidth}
    \includegraphics[width=\linewidth]{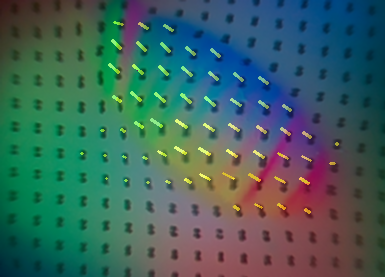}
    \caption{}
  \end{subfigure}

  \caption{(a) Simple 2D relative-motion tracking from motion field measurements. (b) Marker motions measured with GelSlim 3.0.}
  \label{fig:2D_tracking}
\end{figure}

As shown in \figref{fig:2D_tracking}, distributed tactile sensor can measure the 3D motion of the sensing element that follows rigid-body motion of the grasped object. 
Suppose the 3D positions of $m$ markers inside contact patch at the initial frame is $(x_0^i,y_0^i,z_0^i)^T$, where $i=1,...,m$. The 3D position of $m$ markers inside contact patch from the $k$-th frame image is $(x_k^i,y_k^i,z_k^i)^T$. Rewrite the position list of markers in contact patch at $k$-th frame in a compact $m \times 3$ point list $\bm{A_k}$.
Then the rigid-body transform from $0$-th to $k$-th frame is rotation matrix $\bm{R_k}$ along with translation vector $\bm{t_k}$. 


Then, the relative-motion of object from $0$-th to $k$-th frame can be estimated by finding the optimal $\bm{R_k}$ and $\bm{t_k}$ so as to:

\begin{equation}
    \min_{\bm{R_k},\bm{t_k}} [(\bm{R_k  A_0+t_k}) -\bm{A_k}] .
    \label{eq:pose tracking}
\end{equation}

This is a typical Least-Squares Fitting problem between two 3-D point sets and can be solved efficiently using Singular Value Decomposition (SVD) \cite{SArun1987}. 

\begin{figure}[h]
\centering
  \includegraphics[width=\linewidth]{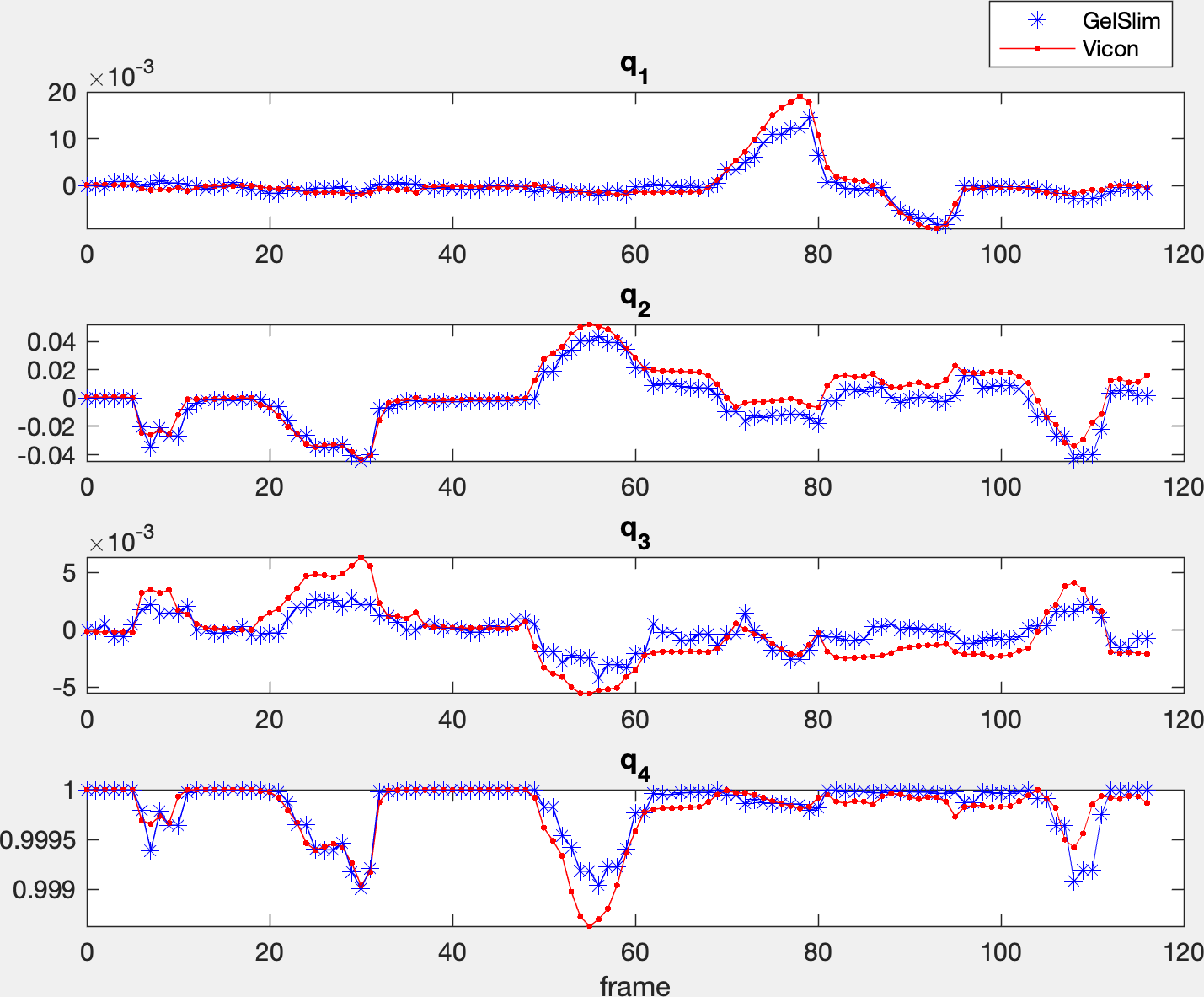}

  \caption{Comparison of tracked relative rotation, represented as quaternion , between GelSlim 3.0 (blue) and Vicon (red, as groundtruth). }
  \label{fig:gelslim_vs_vicon}
\end{figure}

We compared the tactile-measured relative motion with Vicon-tracked relative motion (serving as ground truth). Fig.\ref{fig:gelslim_vs_vicon} shows a comparison of retrieved quaternion from both channels and the good matching indicates that tactile tactile sensor can measure small relative-motion at a precision close to that of Vicon.

\section{Case Study: Fixed Point and Line Contacts}
\label{sec:case_study}

\begin{figure}[t]
\centering
  \includegraphics[width=0.5\linewidth]{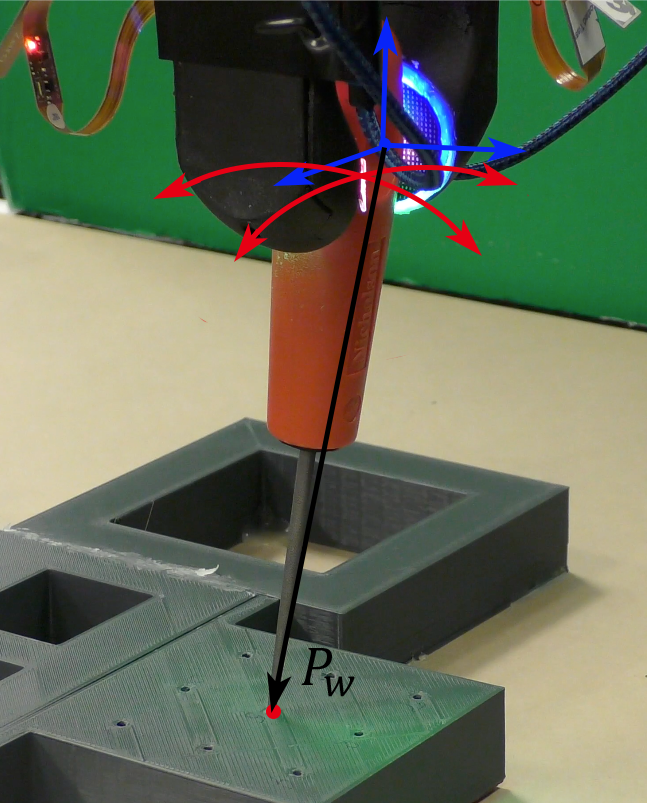}
  \caption{Experiment setup of estimating the Fixed-Point Contact between a grasped object (screwdriver) and the environment.}
  \label{fig:point_contact_setup}
\end{figure}

We validate the proposed framework of localizing extrinsic contact with two case studies: 1) a  fixed point contact, as shown in \figref{fig:point_contact_setup}. 2) a line contact between the moving box and fixed an environment edge, as shown in \figref{fig:simulation}.

\subsection{Fixed Point Contact}
\subsubsection{Constraint Formulation}
As illustrated in chapter II, given sequential measurements of relative motion of the object \{$\bm{R_k}$,$\bm{t_k}$\} ($k=1,2,...,N$), we regress 
\begin{equation}
\min_{\bm{P_b}} \sum_{k=1}^{N} [||(\bm {R_k} \bm {P_b} + \bm {t_k}) - \bm {P_b}||]^2  \label{eq:point_contact_constraint}
\end{equation}
to find the location of the fixed point, $P_b$, in the grasped-object's body frame.

This optimization can be seen as finding the best solution for a stack of the following constraints:

\begin{equation}
(\bm {I}(3) - \bm{R_k ) P_b} = \bm {t_k}  
\label{eq:point_contact_rewrite}
\end{equation}
The best estimation is $\bm{P_b} = (\bm{A}^T\bm{A})^{-1}(\bm{A}^T \bm{r})$, where $\bm{A}$ is a stacked matrix of $(\bm{I}(3)-\bm{R_k})$ and $\bm{r}$ is a stacked vector of $\bm{t_k}$ over the time span.

Eqn.\ref{eq:point_contact_rewrite} is ill-posed and close to singularity when relative rotation matrix $\bm{R}(t)$ is close to $\bm{I}(3)$. Intuitively, good accuracy of a point contact estimation relies on the magnitude of the measured rotation.

\subsubsection{Experiment validation}
\begin{figure}[h]
\centering
  \includegraphics[width=0.65\linewidth]{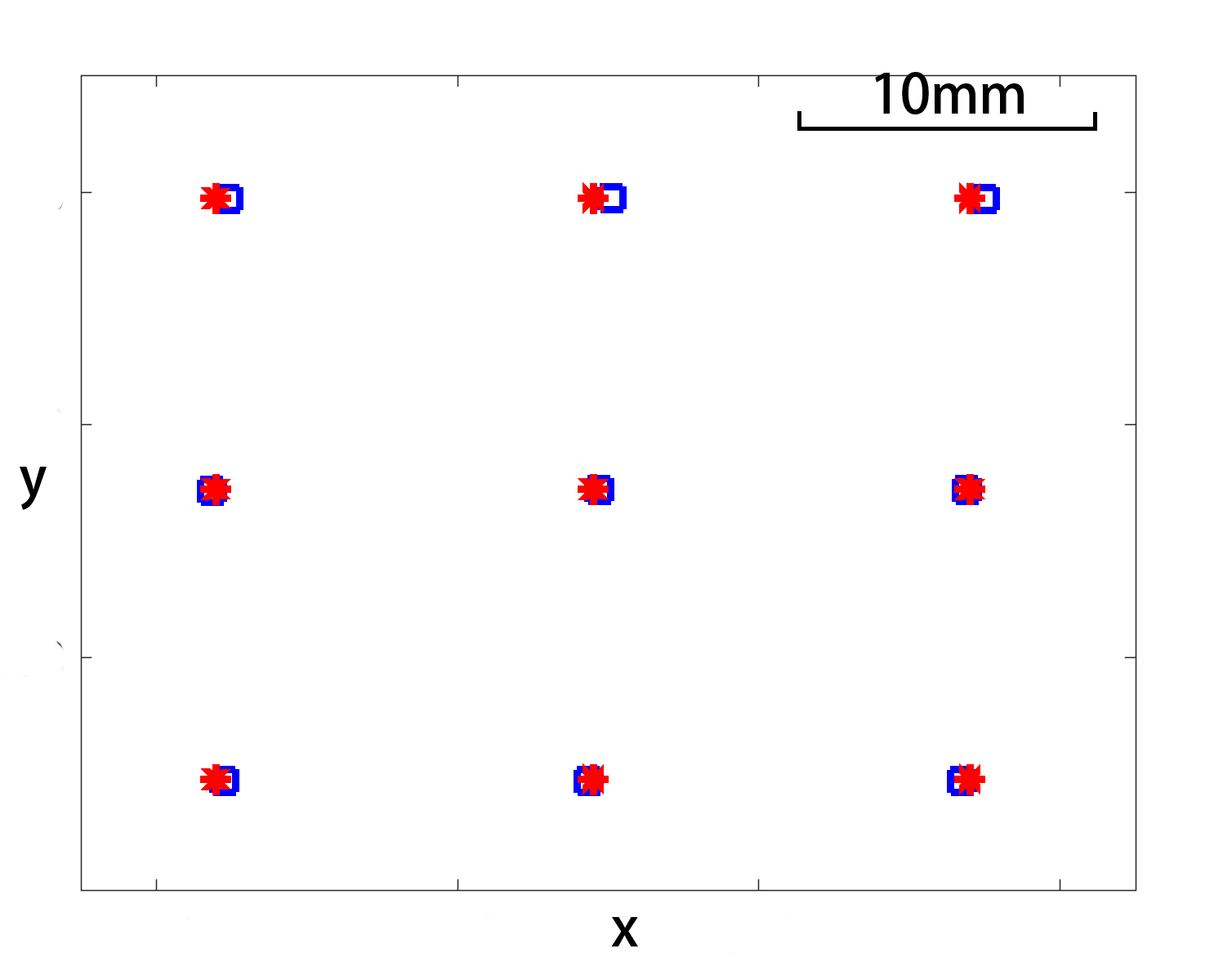}
  \caption{Comparison of the estimated point contact locations (blue square) \textit{v.s.} ground truth (red star).}
  \label{fig:Comparison_point_contact}
\end{figure}

We carried out an experiment to validate localization of a fixed point. The experiment setup is shown in Fig.\ref{fig:point_contact_setup}. A robot gripper, equipped with GelSlim sensors,  grasps a screwdriver whose tip is trapped in one of the holes on a flat plate. The robot grasps the screwdriver and executes small motions to explore the motion space of the screwdriver. The tactile sensors measure relative motion of the screwdriver. We retrieve the point location using the proposed extrinsic contact sensing framework and the comparison of estimated results with groudtruth is shown in Fig.\ref{fig:Comparison_point_contact} 

\subsection{Line Contact}
This case studies the localization of an edge in the environment that is in contact with an moving box. If no slip is allowed, then this localization corresponds to localizing by solving both a fixed point constraint (eqn.\ref{eq:fixed_point_constraint}) and a fixed direction constraint (eqn.\ref{eq:fixed_direction_constraint}), which is a linear regression that has a single optimal solution. We skip it in this paper and introduce the formulation of a more complex situation, where the box and the edge can slip.

\subsubsection{Constraint Formulation}
Suppose initially the object's bottom surface is on a plane with equation 

\begin{equation}
    \bm{n}_{\bm{0}}\cdot (\bm{x-x_0})=0 
    \textbf{    or    } 
    \bm{n}_{\bm{0}} \cdot \bm{x}=c_0,
\label{eq:bottom surface ini}
\end{equation}

where $\bm{n}_{\bm{0}}$ is the normal direction of the bottom plane and $\bm{x_0}$ is a point in the plane. Then, with the tracked relative motion $\bm{R_k}$ and $\bm{t_k}$, we know that the normal direction of the bottom surface can be written as $\bm{n}_{\bm{k}}=\bm{R}_{\bm{k}}  {\bm{n}_0}$ and the point on bottom surface is now at $\bm{x_k} = \bm{R_k} {\bm{x_0}}+\bm{t_k}$. The equation of bottom surface at the $k$-th frame is thus expressed as:
\begin{equation}
    \bm{n}_{\bm{k}} \cdot (\bm{x-x_k})=0
    \textbf{    or    }     
    \bm{n}_{\bm{k}} \cdot \bm{x}=c_k,
\end{equation}
where $c_k=\bm{n}_{\bm{k}} \cdot  \bm{x}_{\bm{k}}$.

Because of the non-penetration constraints between the object (box) and the edge, the contact edge should be the intersection of the bottom surfaces from all time steps. Thus, the line should be perpendicular to the normal directions of the bottom surfaces for all time steps. 
Therefore, the unit direction vector, $\bm{l}$, of the edge can be regressed by solving the optimization problem in \eqref{eq:contact localization 1}: 
\begin{equation}
    \min_{\bm{l,n_0}} \sum_{k=1}^{n}{( \bm{n}_{\bm{k}} \cdot \bm{l})^2} 
    \label{eq:contact localization 1}
\end{equation}
constrained to $ \bm{l}\cdot  \bm{l}=1$.
Note that Eqn.\ref{eq:contact localization 1} is a nonlinear optimization problem now and numerical solution could be sensitive to the initial guess. 

To determine the position of the edge, a point on the edge should also be decided. Denote the point as $\bm{x_0}$. Since the point $\bm{x_0}$ should be on the intersection of all bottom surfaces, it can be found by solving the optimization problem in \eqref{eq:contact localization 2}:

\begin{equation}
    \min_{\bm{x_0}} \sum_{k=1}^{n}{( \bm{n}_{\bm{k}} \cdot ( \bm{R}_{\bm{k}} \cdot \bm{x_0} + \bm{t}_{\bm{k}} - \bm{x_0}))^2} 
    \label{eq:contact localization 2}
\end{equation}

\subsubsection{Simulation Validation}

We first validate this case with synthetic simulation data. 
\begin{figure}[t]
\centering
  \includegraphics[width=0.45\linewidth]{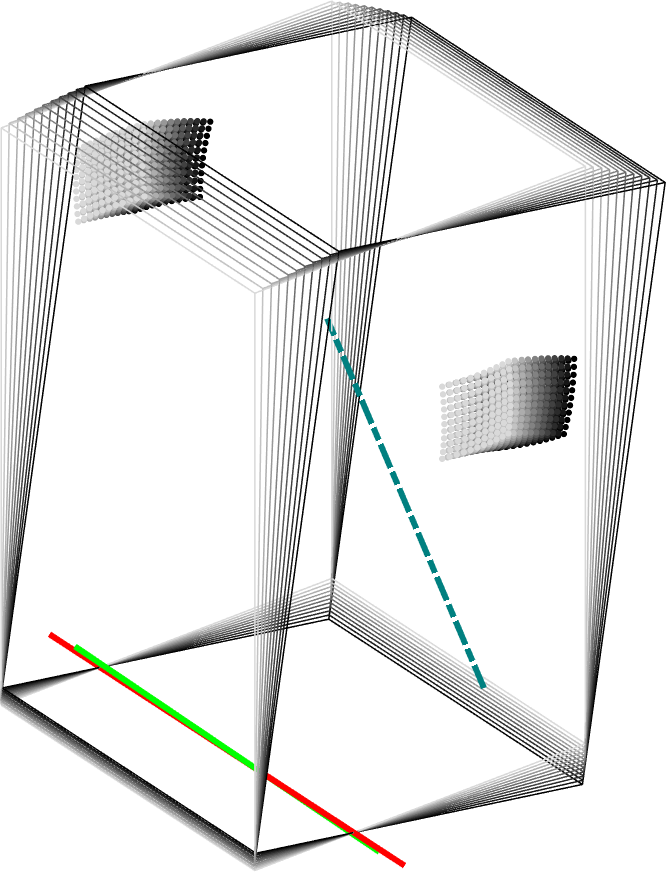}
  \caption{Simulation of extrinsic contact sensing using tactile measurement. Red line represents the ground truth, green line is the estimation, while the cyan dashed line is the initialization input for optimization.}
  \label{fig:simulation}
\end{figure}


A box is set to be in contact with an environment edge, illustrated as a red line in Fig.\ref{fig:simulation}. The box rotates and slides along the edge during the interaction. A virtual tactile sensor is placed on one side of the box and measures 3D motion of  $11\times11$ points on the object with added random noise \arnote{How much noise do we add?}. 
We first solve the optimization problem in \eqref{eq:pose tracking} to estimate the relative pose changes and then we solve the optimization problems in \eqref{eq:contact localization 1} and \eqref{eq:contact localization 2} to estimate the direction and location of the edge.

The simulation results of relative-motion tracking show that the estimated rotation matrices and translation vectors match the ground truth well. The green line in \figref{fig:simulation} represents the estimated edge with estimated pose-change sequence shown with grey boxes. The red line represents the ground truth position of edge, which shows very good alignment with the ground truth. 


\subsubsection{Real Experimental Validation}
We carried out experiments for the fixed-line contact case.
We achieved qualitatively good estimation results for those experiments where the line contact was maintained during the passive exploration. However, we observed that in most cases, in the absence of any controller that enforces it, the line contact is quite fragile, and easily degenerates into a point contact. As a result, many experiments for line contact estimation yield data that violates the assumptions baked into the estimation framework, and lead to poor results. A stronger implementation of this method would require either a more general estimation framework that considers multiple hypothesis for contact formations or a controller that maintains one. 





\section{Discussion}

In this paper we introduced the notion of extrinsic contact sensing, by which we exploit distributed tactile measurements to estimate external contacts on the grasped objects without knowing their geometry. It provides  an  estimation framework  for contact  localization  with  potential  impact  in  contact-rich  manipulation  scenarios. We would like to highlight the value of distributed tactile measurements along three areas:


\subsection{Distributed Tactile Sensing \textit{v.s.} Aggregated F-T Sensing}


Manipulation of a rigid object relies on knowing both the motion of the object and the forces that act on it. Force torque sensors measure force/torque resultants, while distributed tactile sensors can provide 3D force distributions which provide richer detail of the interaction. However, this paper shows that tactile sensors can also provide an important feature, local motion field at contact, that is critical for closed-loop manipulation. 

The strain field in tactile sensing contains multi-modal information about the interaction. First, we can estimate the distribution of contact force~\cite{ma2019dense}. 
Second, the measurement of the motion field from distributed tactile sensors presents a coordinated point cloud that relates the object's rigid-body motion, which also encodes information about the kinematic and friction constraints imposed by the external contacts. Such a relative finite motion, which cannot be perceived from FT sensor resultants, is key to extrinsic contact sensing.


\subsection{Extrinsic Contact Sensing}
An important feature of the proposed framework is that it can work on objects of unknown geometry. The tracking of local relative displacements of the object doesn't need any information about the geometry of the object, while the constraint formulation may only need knowledge of local geometry around the contact. Therefore, it enables estimation on unknown objects with unknown grasps. It's also worth noting that, the object relative-motion is described in the sensors' base frame that is not necessarily calibrated according to the world frame. Therefore, localization of the contact is estimated in the sensor frame, making it a good match for gripper-centered control loop. 

Even though we assume the environment to be fixed in space, this  'environment' could also be another object that is grasped and tracked by another robot, for example in the scenario of assembling with a two-handed robot or multi-robot collaboration.


\subsection{Contact Type Characterization and Active Exploration} %
The key limitation to this work lies in having to assume a particular contact type to formulate the constraints. A better system would need to be able to distinguish contact types.

Motions induced by passive exploration usually result in little rotation or poor estimation accuracy. Since the motion space from a more constrained contact can be a subspace of that from a less constrained one, there can also be ambiguity in contact type estimation if kinematic-constraint-space exploration is not comprehensive. In order to remove the ambiguity, it's vital to guarantee that all constraints to object's motion are fully explored.  Therefore, a more general version of extrinsic contact sensing would benefit from a active exploration process that explores all the allowed motion space dimensions with larger rotation near the current configuration by generating motions corresponding to the estimated contact type and location, and keeping track of when contact formations are maintained or broken. The active exploration calls for a close-loop tactile controller.



\bibliographystyle{bibliographies/IEEEtran} 
{\footnotesize \bibliography{main}} 

\end{document}